\documentclass[11pt]{article}
\usepackage{dialogue2021}

\usepackage{ifxetex}
\ifxetex
    \usepackage{fontspec}
    \setromanfont{Times New Roman}
\else
  \usepackage[T1]{fontenc}
  \usepackage[utf8]{inputenc}
  \usepackage{cmap}
  \usepackage{times}
  \usepackage{latexsym}

\fi

\usepackage[russian,british]{babel}
\usepackage{url}
\usepackage{pgf}

\usepackage{covington} 

\dialogfinalcopy 

\title{Methods for Detoxification of Texts for the Russian Language}

\author{Daryna Dementieva$^\ddag$, Daniil Moskovskiy$^\ddag$, Varvara Logacheva$^\ddag$, David Dale$^\ddag$, \\ \textbf{Olga Kozlova}$^\dag$, \textbf{Nikita Semenov}$^\dag$, and \textbf{Alexander Panchenko}$^\ddag$ \\
$^\ddag$Skolkovo Institute of Science and Technology, Moscow, Russia \\
$^\dag$Mobile TeleSystems (MTS), Moscow, Russia \\
{\{daryna.dementieva, daniil.moskovskiy, v.logacheva, d.dale, a.panchenko\}@skoltech.ru} \\
{\{oskozlo9,nikita.semenov\}@mts.ru}
}

\date{}

\begin{document}
\maketitle
\begin{abstract}
    We introduce the first study of automatic detoxification of Russian texts to combat offensive language. Such a kind of textual style transfer can be used, for instance, for processing toxic content in social media. While much work has been done for the English language in this field, it has never been solved for the Russian language yet. We test two types of models -- unsupervised approach based on BERT architecture that performs local corrections and supervised approach based on pretrained language GPT-2 model -- and compare them with several baselines. In addition, we describe evaluation setup providing training datasets and metrics for automatic evaluation. The results show that the tested approaches can be successfully used for detoxification, although there is room for improvement.
  
  \textbf{Keywords:} text style transfer, toxicity detection, detoxification, pre-trained models
  
\end{abstract}

\selectlanguage{russian}
\begin{center}
  \russiantitle{Методы детоксификации текстов для русского языка}

\medskip
Дарина Дементьева$^\ddag$, Даниил Московский$^\ddag$, Варвара Логачева$^\ddag$, Давид Дале$^\ddag$, \\ Ольга Козлова$^\dag$, \textbf{Никита Семенов}$^\dag$, \textbf{Александр Панченко}$^\ddag$ \\
$^\ddag$Сколковский Институт Науки и Технологий, Москва, Россия \\
$^\dag$Мобильные ТелеСистемы (МТС), Москва, Россия \\
{\{daryna.dementieva, daniil.moskovskiy, v.logacheva, d.dale, a.panchenko\}@skoltech.ru} \\
{\{oskozlo9,nikita.semenov\}@mts.ru}
\medskip

\end{center}

\begin{abstract}
  Мы представляем первое в своем роде исследование автоматической детоксикации русскоязычных текстов для борьбы с оскорбительной речью. Такой перенос стиля для текстов может быть использован, например, для предварительной обработки в социальных сетях. В то время как решения подобных задач уже были преставлены для английского языка, для русского такая постановка задачи и методы её решения описываются впервые. Мы провели эксперименты по тестированию двух типов моделей -- метод обучения без учителя на основе архитектуры BERT, который выполняет локальные коррекции, и метод обучения с учителем на основе предобученой языковой модели GPT-2 -- и сравнили их с несколькими базовыми подходами. Кроме того, мы предоставили описание методологии оценки вместе с набором обучающих данных и метрик для автоматической оценки.  Результаты показали, что протестированные методы могут быть успешно использованы для детоксикации, однако могут быть усовершенствованы. 
  
  \textbf{Ключевые слова:} перенос стиля для текстов, определение токсичности, детоксификация, предобученные модели 
\end{abstract}
\selectlanguage{british}

\section{Introduction}
Global access to the Internet has enabled the spread of information all over the world and has given many new possibilities. 
On the other hand, alongside the advantages, the exponential and uncontrolled growth of user-generated content on the Internet has also facilitated the spread of toxicity and hate speech. 
Much work has been done in the direction of offensive speech detection \cite{d2020towards, schmidt2017survey,pamungkas-patti-2019-cross}. However, it has become essential not only to detect toxic content but also to combat it in smarter ways. While some social networks block sensitive content, another solution can be to detect toxicity in a text which is being typed in and offer a user a non-offensive version of this text. 
This task can be considered a style transfer task, where the source style is toxic, and the target style is neutral/non-toxic.

The task of style transfer is the task of transforming a text so that its content and the majority of properties stay the same, and one particular attribute (\textit{style}) changes. This attribute can be the sentiment~\cite{shen2017style, melnyk2017improved}, the presence of bias~\cite{pryzant2020automatically}, the degree of formality~\cite{rao2018dear}, etc. The work~\cite{jin2020deep} gives more examples of style transfer applications. 
Considering the task of detoxification, it has already been tackled by different groups of researchers~\cite{dos2018fighting,tran2020towards}, as well as a similar task of transforming text to a more polite form~\cite{madaan2020politeness}. However, all these works deal only with the English language.  
As for Russian, the methods of text style transfer and text detoxification have not been explored before. 

To the best of our knowledge, our work is the first effort to solve the text style transfer task with a focus on toxicity elimination for the Russian language. 
We leverage pre-trained language models (GPT and BERT) and demonstrate that they can successfully solve the task after being trained on a very small parallel corpus or only on non-parallel data.

The contributions of this work are three-fold:
\begin{enumerate}
    \item We introduce the new study of text detoxification for the Russian language;
    \item We conduct experiments with two well-performing style transfer methods: a method based on \mbox{GPT-2} which rewrites the text and a BERT-based model which performs targeted corrections;
    \item We create an evaluation setup for the style transfer task for Russian: we prepare the training and the test datasets and implement two baselines.
\end{enumerate}

\section{Problem Statement}
The definition of \textit{textual style} in the context of NLP is still vague~\cite{tikhonov2018wrong}. One of the first definitions of style refers to how the sense is expressed~\cite{mcdonald1985computational}. However, in our work, we adhere to the data-driven definition of style. Thus, the style simply refers to the characteristics of a given corpus that are distinct from a general text corpus~\cite{jin2020deep}. The style is a particular characteristic from a set of categorical values: \{\texttt{positive}, \texttt{negative}\} \cite{shen2017style}, \{\texttt{polite}, \texttt{impolite}\} \cite{madaan2020politeness}, \{\texttt{formal}, \texttt{informal}\} \cite{rao2018dear}. Commonly, it is assumed that this textual characteristic is measurable using a function $g(x_i) \rightarrow s_i$ that gets as input text $x_i$ and returns the corresponding style label $s_i$. For instance, it can be implemented using a text classifier.

We define the task of style transfer as follows. Let us consider two corpora ${D^X}=\{x_1, x_2, ..., x_n\}$ and ${D^Y} = \{y_1, y_2, ..., y_m\}$ in two different styles -- $s^X$ and $s^Y$, respectively. The task is to create a model $f_\theta: X \rightarrow Y$, where $X$ and $Y$ are all possible texts with styles $s^X$ and $s^Y$. The task of selecting the optimal set of parameters $\theta$ for $f$ consists maximising the probability $p(y'|x, s^Y)$ of transferring a sentence $x$ with the style $s^X$ to the sentence $y'$ which saves the content of $x$ and has the style $s^Y$. The parameters are maximised on the corpora $D^X$ and $D^Y$ which can be parallel or non-parallel. We focus on the transfer $s^X \rightarrow s^Y$, where $s^X$ is the toxic style, and $s^Y$ is neutral.

\section{Related Work}
Style transfer was first proposed and widely explored for images~\cite{gatys2016image}. However, the task of text style transfer has currently gained less attention, partly due to the ambiguity of the term ``style'' for texts. 
Nevertheless, there exists a large body of work on textual style transfer for different styles. All the existing methods can be divided into techniques that use parallel training corpora and those using only non-parallel data. 
The latter category is larger because pairs of texts which share the content but have different styles are usually not available. At the same time, it is relatively easy to find non-parallel texts of the same domain with different styles (e.g. positive and negative movie reviews, speeches by politicians from different parties, etc.).

One of the methods which uses only non-parallel data is \textit{Delete, Retrieve, Generate} \cite{li2018delete} model. It is based on the idea that words in a sentence can be divided into those responsible for the sentence semantics and those carrying the style information. Therefore, if we delete the style words and replace them with the corresponding words of the opposite style, we can change the style of the sentence while keeping the content intact. Alternative to this approach are methods that create disentangled representations of text \cite{john2019disentangled}. In this case, the style and the content of a text are encoded into different spaces. When generating a text with a new style, we substitute the vector of the text style with the vector representation of the target style and generate a new sequence.

On the other hand, if there exists a corpus with parallel sentences $\{(x_1, y_1), (x_2, y_2), ..., (x_n, y_n)\}$ 
then style transfer can be formulated as a sequence-to-sequence task, analogously to supervised Machine Translation, summarization, paraphrasing, etc. Such models can greatly benefit from pre-trained language models, such as GPT~\cite{gpt2} or T5 \cite{raffel2020exploring}. They often perform well on a range of NLP tasks with no fine-tuning. Moreover, when a small training dataset is available, their performance improves even further. For example, in \cite{DBLP:conf/emnlp/KrishnaWI20} a GPT-based model was fine-tuned on an automatically generated parallel corpus to transfer between multiple styles. The recently released ruGPT3\footnote{https://github.com/sberbank-ai/ru-gpts} model allows us to leverage big textual data for the detoxification task in Russian. 

\section{Methodology}

We suggest several solutions to the text detoxification task. We test a method based on the GPT model, which uses parallel data and a BERT-based solution trained solely on non-parallel corpora. We also implement several baselines.

\subsection{Baselines}

\paragraph{Duplicate} This is a naive baseline that amounts to performing no changes to the input sentence. It represents a lower bound of the performance of style transfer models, i.e. it helps us check that the models do not contaminate the original sentence.

\paragraph{Delete} This method eliminates toxic words based on a predefined toxic words vocabulary. The idea is often used on television and other media: rude words are bleeped out or hidden with special characters (usually an asterisk). The main limitation of this method is vocabulary incompleteness: we cannot collect all the rude and toxic words. Moreover, new offensive words and phrases can appear in the language that can be also concatenated with different prefixes and suffixes. 
On the other hand, this method can preserve the content quite well, except for the cases when toxic words contain meaning that is essential for the understanding of the whole text.

\paragraph{Retrieve} This method introduced in \cite{li2018delete} is targeted at improving the accuracy of style transfer. For a given toxic sentence, we retrieve the most similar non-toxic text from a corpus of non-toxic samples. In this case, we get a safe sentence. However, the preservation of the content depends on the corpus size and is likely to be very low.

\subsection{detoxGPT}
\label{sec:ruGPT}
\begin{figure}[h!]
        \centering
        \includegraphics[width=\textwidth]{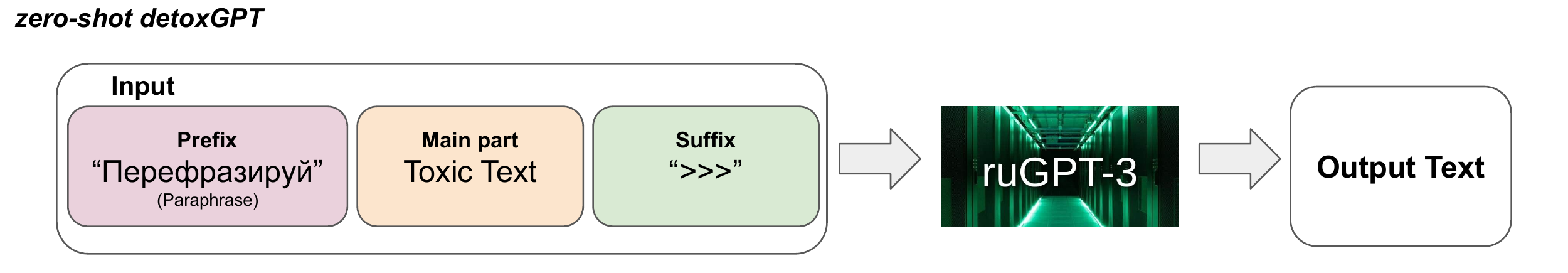}
        \caption{The illustration of pipeline of \textit{zero-shot} setup of detoxGPT approach.}
        \label{fig:detoxGPT_zero_shot}
    \end{figure}

\begin{figure}[h!]
        \centering
        \includegraphics[width=\textwidth]{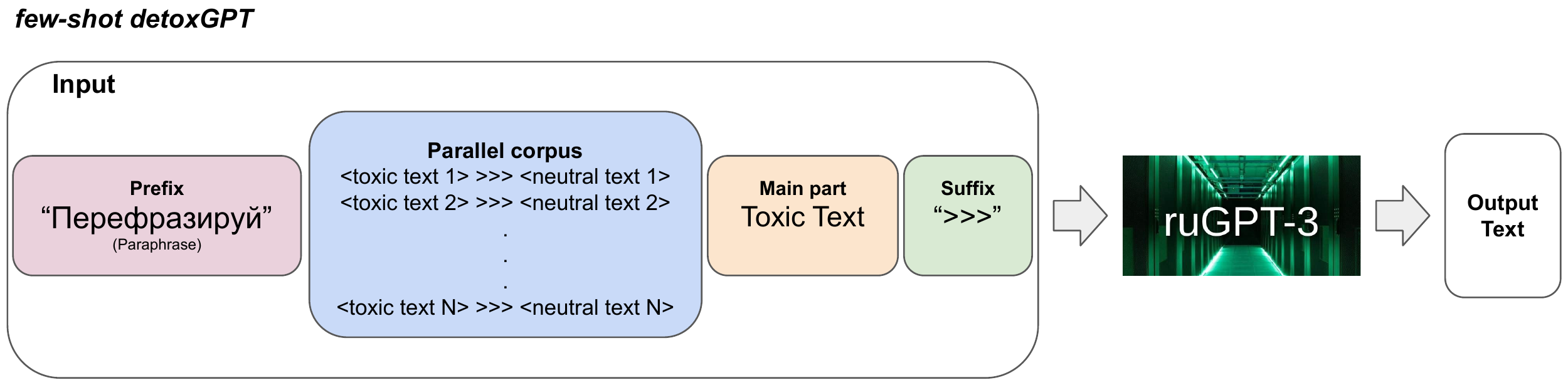}
        \caption{The illustration of pipeline of \textit{few-shot} setup of detoxGPT approach.}
        \label{fig:detoxGPT_few_shot}
    \end{figure}

\begin{figure}[h!]
        \centering
        \includegraphics[width=\textwidth]{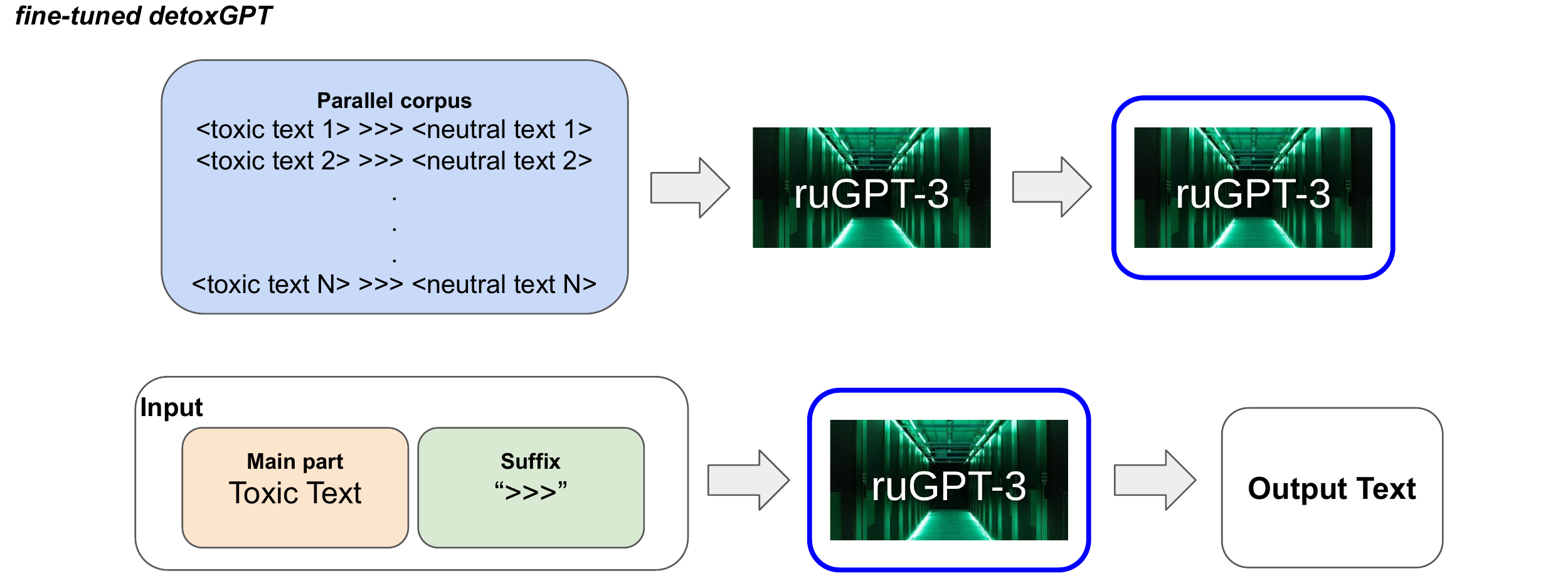}
        \caption{The illustration of pipeline of \textit{fine-tuned} setup of detoxGPT approach.}
        \label{fig:detoxGPT_fine_tuned}
    \end{figure}

GPT-2 \cite{gpt2} is a powerful language model which can be adapted to a wide range of NLP tasks using a very small task-specific dataset. Until recently, there were no such models for Russian. The AI Journey competition\footnote{https://ai-journey.ru} released the ruGPT3 model capable of generating coherent and sensible texts in Russian. 
We suggest using it for style transfer via the following setups:
\begin{itemize}
    \item \textit{\textbf{zero-shot}}: the model is taken as is (with no fine-tuning). The input is a toxic sentence which we would like to detoxify prepended with the prefix ``\foreignlanguage{russian}{Перефразируй}'' (rus. \textit{Paraphrase}) and followed with the suffix ``$>>>$'' to indicate the paraphrasing task. ruGPT3 has already been trained for this task, so this scenario is analogous to performing paraphrasing. The schematic pipeline of this setup is presented in Figure \ref{fig:detoxGPT_zero_shot}.
 
   \item \textit{\textbf{few-shot}}: the model is taken as is. Unlike the previous scenario, we give a prefix consisting of a parallel dataset $\{(t_1^X, t_1^Y), ..., (t_n^X, t_n^Y)\}$ of toxic and neutral sentences in the following form: ``$t_i^X >>> t_i^Y$''. These examples can help the model understand that we require \textit{detoxifying} paraphrasing. The parallel sentences are followed with the input sentence which we would like to detoxify with the prefix ``\foreignlanguage{russian}{Перефразируй}'' and the suffix $>>>$. The schematic pipeline of this setup is presented in Figure \ref{fig:detoxGPT_few_shot}.

    \item \textit{\textbf{fine-tuned}}: the model is fine-tuned for the paraphrasing task on a parallel dataset $\{(t_1^X, t_1^Y), ..., (t_n^X, t_n^Y)\}$. This implies training of the model on strings of the form ``$t_i^X >>> t_i^Y$''. After the training, we give the input to the model analogously to the other scenarios. The schematic pipeline of this setup is presented in Figure \ref{fig:detoxGPT_fine_tuned}.
 \end{itemize}

The described methods require parallel data. These have to be pairs of sentences with the same content and the different toxicity level. Such sentences are not created ``naturally'' (unlike translations of the same text into different languages), so they have to be written from scratch to train such models. This is a laborious process. However, our intuition is that the detoxGPT model can perform detoxification after being trained on a very small number (several hundred) of parallel sentences, which can be created quickly. 

\subsection{condBERT}
\label{section:condbert}

BERT (Bidirectional Encoder Representations from Transformers) \cite{devlin2019bert} is a masked language model which has been trained on the task of predicting a missing word given the rest of the sentence. Although BERT is mainly used for getting word vector representations or sequence labeling and text classification tasks, it can also be used in the gap-filling scenario, i.e. for retrieving a word in a context that has been replaced with a [MASK] token. This scenario perfectly suits the delete-retrieve-generate style transfer method, which replaces individual words of a sentence and, as a result, generates so-called ``lexical substitution" \cite{arefyev2020always}. 

To make BERT fully suitable for style transfer, we need to change the model so that masking and replacing words changes the style of the input sentence. This can be done via fine-tuning BERT on style-specific corpora for the source and the target styles so that it learns the word distributions conditioned on a style and makes replacements that agree with it. Such a BERT-based model was first applied to the data augmentation task in \cite{wu2019conditional}. Then, in \cite{wu2019mask}, a similar model was used for sentiment style transfer.

The model \textbf{condBERT} (conditional BERT) model was proposed in \cite{wu2019conditional}. While the tokens to replace were selected randomly in the original work, we mask tokens associated with the source style (toxic). To select the toxic words, we train a bag-of-words logistic regression model, which classifies the sentences as toxic or neutral. As a by-product of this model, we acquire weights for each word from the vocabulary. These weights can be interpreted as the toxicity level. We consider a token to be toxic if its weight is higher than a predefined threshold.

\begin{figure}[h!]
        \centering
        \includegraphics[scale=0.65]{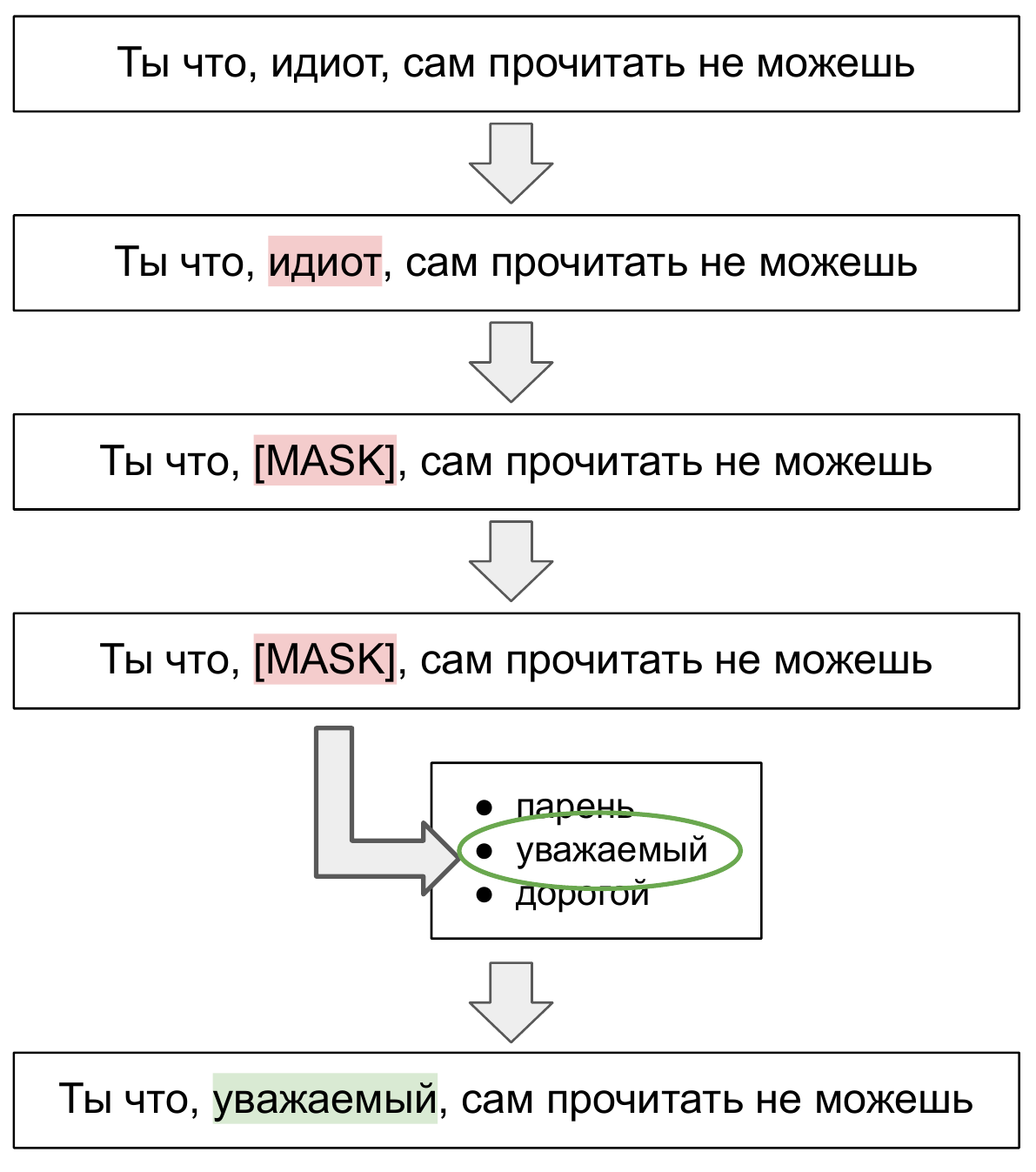}
        \caption{The illustration of the main idea of condBERT approach.}
        \label{fig:condBERT}
    \end{figure}

We then train the model on two corpora $D^X$ and $D^Y$ for the source and the target styles. To teach the model to distinguish styles, we include the style information as an extra embedding layer as described in~\cite{wu2019conditional}. Thus, it learns different distributions for toxic and non-toxic texts. To further force the model to replace toxic tokens with tokens that have a close meaning and are not toxic, we calculate the toxicity level of each token in the BERT vocabulary (using the logreg weights) and penalize the predicted probabilities of tokens that have a high toxicity. Finally, we enable condBERT to replace a single [MASK] token with multiple words. We generate the next tokens progressively by beam search and score each multitoken sequence by the harmonic mean of the probabilities of its tokens. The schematic illustration of condBERT approach is presented in Figure \ref{fig:condBERT}.

To evaluate the efficiency of BERT fine-tuning, we test condBERT in two scenarios:
\begin{itemize}
    \item \textit{\textbf{zero-shot}} where BERT is taken as is (with no extra fine-tuning);
    \item \textit{\textbf{fine-tuned}} where BERT is fine-tuned on a dataset of toxic and safe sentences to acquire a style-dependent distribution, as described above.
\end{itemize}

The scenarios are different only in terms of BERT pre-training. 
They both use the classifier-based selection of toxic words and penalties for the toxicity of word replacements.

The strength of condBERT compared to the GPT-based method is that it does not require any parallel data. Besides that, it does not rewrite the sentence, which might be a better strategy in terms of content preservation.

\section{Evaluation}
To perform a comprehensive evaluation of a style transfer model, we need to make sure that it (i) changes the text style, (ii)~preserves the content, and (iii)~yields a grammatical sentence. The majority of works on style transfer use individual metrics to evaluate the three parameters. 
However, \cite{pang2019unsupervised} points out that these three parameters are usually inversely correlated, so they need to be combined to find the balance. Our evaluation setup (individual metrics and the joint metric which combines them) follows this principle.

\subsection{Style transfer accuracy} To evaluate style transfer accuracy (\textbf{STA}), we train a binary classifier $g(x_i) \rightarrow s_i$ based on RuBERT \cite{kuratov2019adaptation} that classifies text $x_i$ into style $s_i \in$\{\texttt{toxic}, \texttt{neutral}\}. We fine-tune the RuBERT model on RuToxic dataset (see Section \ref{section:dataset}). It achieves the F$_1$-score of 0.83 on a held-out test set. Thus, it shows a reasonable result on detection of toxic texts and can be used for evaluating the strength of style transfer. Since we want to perform the detoxification task, we expected the outputs of style transfer methods to be non-toxic. 
We compute the accuracy based on this assumption. 

\subsection{Content preservation} We approach the assessment of content preservation from two sides. First, we calculate word-based metrics: (i)~the unigram word overlap (\textbf{WO}) between the tokens of the original sentence $x$ and the style-transferred sentence $y$: $\frac{count(x \cap y)}{count(x \cup y)}$ and (ii)~\textbf{BLEU} score, which is the ngram precision for $n$ from 1 to 4. Secondly, we calculate the cosine similarity (\textbf{CS}) between the vector representations of the input and the output sentences. We calculate vector representations as the mean of token vector representations extracted with a fastText \cite{DBLP:journals/tacl/BojanowskiGJM17} model from RusVectores\cite{KutuzovKuzmenko2017}.\footnote{http://vectors.nlpl.eu/repository/20/213.zip}

\subsection{Language quality} We use perplexity (\textbf{PPL}) to evaluate the quality of the generated sentence. As a language model for this metric, we use the ruGPT2Large\footnote{https://github.com/sberbank-ai/ru-gpts\#Pretraining-ruGPT2Large} model which was trained on bigger amount of content than used ruGPT3 models and was not used in our detoxGPT setups. Thus, we can claim that this model can give us the fair score for the perplexity.

\subsection{Aggregated metric} Following \cite{pang2019unsupervised}, we combine the three parameters. Namely, we compute the geometric mean of STA, CS, and 1/PPL:
\begin{center}
$
\mbox{GM} = (max(\mbox{STA}, 0) \times max(\mbox{CS}, 0) \times max(1/\mbox{PPL}, 0))^\frac{1}{3}
$
\end{center}
We denote this joint metric as \textbf{GM}. Other content preservation metrics do not participate in the combination and are reported to understand the model properties better.

Although there are still discussions about the efficiency of the usage of automatic metrics for the evaluation \cite{DBLP:journals/corr/abs-2004-05001} of style transfer tasks, we believe that the described metrics can adequately illustrate the strength of style transfer methods.

\section{Experiments}

We train and evaluate the two proposed models (detoxGPT and condBERT) and compare them to the baselines.

\subsection{Datasets}
\label{section:dataset}

All our methods including the \textit{Delete} and \textit{Retrieve} baselines require collections of toxic and non-toxic texts for training. There exist non-parallel corpora of such texts for Russian. Two corpora of toxic comments were released on Kaggle.\footnote{https://www.kaggle.com/blackmoon/russian-language-toxic-comments}$^,$\footnote{ https://www.kaggle.com/alexandersemiletov/toxic-russian-comments} We concatenate these resources and denote the joint corpus \textbf{RuToxic} dataset.
It consists of 163,187 texts (31,407 (19\%) toxic and 131,780 non-toxic) from the Russian social networks Odnoklassniki\footnote{https://ok.ru} and Pikabu.\footnote{https://pikabu.ru}

We also use a fraction of this dataset to construct the parallel training data for detoxGPT: we select 200 toxic sentences and manually rewrite them into non-toxic ones. Besides, we use the RuToxic dataset to train toxicity weights for condBERT.

We test all models on 10,000 randomly selected toxic sentences from RuToxic. These sentences are not used for training.

\subsection{Experimental Setup}

For the \textbf{Delete} method, we use a manually created set of rude, obscene, and toxic words. We extend the list with word lemmas for better coverage. 
In \textbf{Retrieve} method we get the word vector representations from Russian \textit{fasttext} model from the RusVectores website. The text vector representations are obtained as the mean of token vectors. We use cosine similarity as the metric of similarity between texts. For both Delete and Retrieve methods the input was preprocessed with the following steps: the input text was tokenized and obtained tokens were lemmatized with UDPipe. \footnote{https://ufal.mff.cuni.cz/udpipe/1/models} 

\textbf{ruGPT3} model is available in three flavours: \texttt{small} (125m parameters with 2048 context), \texttt{medium} (350m parameters with 2048 context), and \texttt{large} (760m parameters with 2048 context). We experiment with all of them. We denote the detoxGPT models that use these ruGPT3 pretrained LMs as detoxGPT-small, detoxGPT-medium, and detoxGPT-large.
ruGPT3 uses the following hyper-parameters:
\begin{itemize}
    \item \textbf{top\_k}: integer parameter that is greater or equal to 1. Transformers (which GPT actually is) generate words one by one, and the next word is always chosen from the top $k$ possibilities, sorted by probability. We use top\_k~=~3.
    \item \textbf{top\_p}: floating-point parameter from 0 to 1. The idea is similar to the top\_k parameter, but the sampling is done by choosing from the smallest possible set of words whose cumulative probability exceeds the probability $p$. We use top\_p~=~0.95. 
    \item \textbf{temperature} ($t$): floating-point parameter  greater or equal to 0. It represents the degree of freedom for the model. For the higher temperatures (e.g. 100), the model can start a dialogue instead of paraphrasing, whereas for a temperature of around 1 it barely changes the sentence. We use $t$~=~50.
\end{itemize}
For the few-shot and fine-tuned scenarios, we used the dataset with 200 parallel samples as described in Section \ref{section:dataset}.

For \textbf{condBERT} we use two setup of pre-trained weights: 
\begin{itemize}
    \item Conversational RuBERT\footnote{https://huggingface.co/DeepPavlov/rubert-base-cased-conversational} from DeepPavlov \cite{kuratov2019adaptation}; 
    \item A smaller version of multilingual BERT for Russian\footnote{https://huggingface.co/Geotrend/bert-base-ru-cased} 
    from Geotrend \cite{abdaoui-etal-2020-load}.
\end{itemize}
The BERT model from DeepPavlov is more commonly used for Russian language, but it is shipped without the masked LM layer that has to be trained from scratch. The BERT from Geotrend, conversely, has a pretrained LM head.

\subsection{Results and Discussion}

The performance of the proposed models on this data is shown in Table \ref{tab:results}.

\begin{table}[ht!]
    \centering
    \begin{tabular}{|p{3.5cm}|c|c|c|c|c|c|}
        \hline
        Method & STA$\uparrow$ & CS$\uparrow$ & WO$\uparrow$ & BLEU$\uparrow$ & PPL$\downarrow$ & GM$\uparrow$ \\
        \hline
        Duplicate 										& 0.00 & 1.00 & 1.00 & 1.00 & \ 146.00 & 0.05 \ $\pm$ 0.0012 \\
        \hline
        Delete 											& 0.27 & 0.96 & 0.85 & 0.81 & 263.55 & 0.10 \ $\pm$ 0.0007 \\
        Retrieve 										& 0.91 & 0.85 & 0.07 & 0.09 & \ 65.74 & 0.22 \ $\pm$ 0.0010 \\
        \hline
        detoxGPT-small & & & & & & \\
        \multicolumn{1}{|l|}{\hspace{1.25cm}zero-shot} & \textcolor{gray}{0.93} & \textcolor{gray}{0.20} & \textcolor{gray}{0.00} & \textcolor{gray}{0.00} & \textcolor{gray}{159.11} & \textcolor{gray}{0.10 \ $\pm$ 0.0005} \\
        \multicolumn{1}{|l|}{\hspace{1.25cm}few-shot} & 0.17 & 0.70 & 0.05 & 0.06 & \ 83.38 & 0.11 \ $\pm$ 0.0009 \\
        \multicolumn{1}{|l|}{\hspace{1.25cm}fine-tuned} & 0.51 & 0.70 & 0.05 & 0.05 & \ 39.48 & 0.20 \ $\pm$ 0.0011 \\
        detoxGPT-medium & & & & & & \\
        \multicolumn{1}{|l|}{\hspace{1.25cm}fine-tuned} & 0.49 & 0.77 & 0.18 & 0.21 & \ 86.75 & 0.16 \ $\pm$ 0.0009 \\
        detoxGPT-large & & & & & & \\
        \multicolumn{1}{|l|}{\hspace{1.25cm}fine-tuned} & 0.61 & 0.77 & 0.22 & 0.21 & \ \textbf{36.92} & \textbf{0.23}* $\pm$ 0.0010 \\
        \hline
        condBERT & & & & & & \\ 
        \multicolumn{1}{|l|}{\hspace{1.25cm}DeepPavlov zero-shot} & 0.53 & 0.80 & 0.42 & 0.61 & 668.58 & 0.08 \ $\pm$ 0.0006 \\
        \multicolumn{1}{|l|}{\hspace{1.25cm}DeepPavlov fine-tuned} & 0.52 & 0.86 & 0.51 & 0.53 & 246.68 & 0.12 \ $\pm$ 0.0007 \\
        \multicolumn{1}{|l|}{\hspace{1.25cm}Geotrend zero-shot} & 0.62 & 0.85 & 0.54 & \textbf{0.64} & 237.46 & 0.13 \ $\pm$ 0.0009 \\
        \multicolumn{1}{|l|}{\hspace{1.25cm}Geotrend fine-tuned} & \textbf{0.66} & \textbf{0.86} & \textbf{0.54} & 0.64 & 209.95 & 0.14 \ $\pm$ 0.0009 \\
        \hline
    \end{tabular}
    \caption{The results of evaluation of proposed detoxification approaches. \textbf{STA}: Style transfer accuracy. \textbf{CS}: Cosine similarity. \textbf{WO}: Word overlap rate. \textbf{PPL}: Perplexity. \textbf{GM}: Geometric mean. The larger$\uparrow$ (or lower$\downarrow$), the better. \textcolor{gray}{Gray} numbers show that a method singificantly fails to preserve the content. The values \textbf{in bold} are the best scores. The asterisk * denotes the improvement over the \textbf{Retrieve} baseline that is statistically significant at $p \leq 0.01$. The standard deviations of \textbf{GM} are calculated by bootstrapping the test dataset.}
    \label{tab:results}
\end{table}

The baseline approaches represent the two extremes: while \textbf{Delete} gains a low STA and high content similarity, the \textbf{Retrieve} method, on the contrary, achieves a relatively high STA with extremely low WO and BLEU. These results are natural since the Delete method only eliminates toxic words and leaves the rest of the sentence intact, which results in high word-based similarity. At the same time, such deletion of words often ruins the sentence structure and results in high PPL. The Retrieve method always outputs only non-toxic, fully human-readable sentences; this strategy achieves a high STA score and the highest GM score between baselines. However, the content of such sentences is unpredictable and usually very different from the original input.

We experiment with \textit{zero-shot}, \textit{few-shot}, and \textit{fine-tuned} setups for the three \textbf{detoxGPT} model versions as described in Section~\ref{sec:ruGPT}. However, the quality of the output of the \textit{zero-shot} and \textit{few-shot} scenarios is poor for all models. Thus, we report the results of \textit{zero-shot}, \textit{few-shot} only for the detoxGPT-small model to illustrate the difference in scores. Table \ref{tab:results} shows that content similarity and fluency of both \textit{zero-shot} and \textit{few-shot} models are lower than those of the baselines. The \textit{zero-shot} method manages to reach high style accuracy by generating completely irrelevant texts which happen to be mostly non-toxic. As a result, we do not take into account its results in comparison with other approaches. On the other hand, when fine-tuned on only 200 samples, detoxGPT models outperform the baselines. The best results are achieved by the \textbf{detoxGPT-large} model. It reaches the highest values for all metrics (and the lowest for PPL which stands for the highest naturalness) including the joint GM score.

The \textbf{condBERT}-based models also outperform the \textbf{Delete} baseline, but fall short of the \textbf{Retrieve} baseline due to lower fluency. The condBERT models based on Geotrend pre-trained BERT model show better performance than DeepPavlov setup in general. The reason is the pre-trained language model part in Geotrend RuBERT. For DeepPavlov setup these weights of the model were not  pre-trained and were initialized randomly. The comparison with detoxGPT is ambiguous: condBERT attains the highest STA score and larger values of content similarity metrics, while the quality of language is quite low. The high style transfer accuracy is explained by the fact that the replacement words in condBERT are explicitly penalized for toxicity. Nevertheless, the model can generate similar toxic or rude words, making its style transfer quality imperfect. At the same time, as detoxGPT generates the sentences from scratch, we cannot control the content preservation (the models occasionally change it entirely). In contrast, the condBERT model changes only a small fraction of words. However, the downside of condBERT is the loss of naturalness. GPT-based models are free to choose every single word in the output, usually yielding a natural sentence. Conversely, when replacing only single words (as condBERT does), it is difficult to control the overall grammar and fluency. Furthermore, condBERT, for instance, in DeepPavlov setup sometimes substitutes words with special symbols such as [UNK].

\begin{table}[ht!]
\footnotesize
    \centering
    \begin{tabular}{p{3.5cm}|p{11cm}}
        \hline 
        Model & Sentence\\
        \hline 
        Input & \foreignlanguage{russian}{не дай бог моя дочь так оденется убью нахуй палкой} \newline \scriptsize{(If, God forbid, my daughter goes out dressed like this, I'll fucking kill her with a stick)}  \\ [0.1cm]
        Delete & \foreignlanguage{russian}{не дай бог моя дочь так оденется убью палкой} \newline \scriptsize{(If, God forbid, my daughter goes out dressed like this, I'll kill her with a stick)} \\ [0.1cm]
        Retrieve & \foreignlanguage{russian}{не бросайте угла родного одной мы лежали больнице палате в в в те девчонкой была молодой годы} \newline \scriptsize{(don't abandon your native corner same hospital we were ward in in in those girl was young years)} \\ [0.1cm]
        detoxGPT-small & \foreignlanguage{russian}{подскажите, у меня дочь так оденется палкой.} \newline \scriptsize{(Could you give me an advise, my daughter gets dressed as a stick.)} \\ [0.1cm]
        detoxGPT-medium & \foreignlanguage{russian}{да не торопись вы так, бабушка.} \newline \scriptsize{(Granny, don't rush, take your time.}\\ [0.1cm]
        detoxGPT-large & \foreignlanguage{russian}{\underline{\smash{извините, но мне вас очень жаль, если вы начнете так одеваться.}}} \newline \scriptsize{(Excuse me, but I feel sorry for you if you start getting dressed up like this.)}\\ [0.1cm]
        condBERT & \foreignlanguage{russian}{не дай бог моя дочь так оденется станки палкой} \newline \scriptsize{(God forbid, my daughter gets dressed up like this machine stick.)}\\ [0.1cm]
        \hline 
        
        Input & \foreignlanguage{russian}{что ты читала из его произведений? дура.а если не читала,то прочти.} \newline \scriptsize{(Which of his works have you read? You are an idiot. If you haven't read any, go read.)} \\ [0.1cm]
        Delete & \foreignlanguage{russian}{что ты читала из его произведений дура.а если не читала,то прочти} \newline \scriptsize{(Which of his works have you read you idiot. If you haven't read any, go read.)} \\ [0.1cm]
        Retrieve & \foreignlanguage{russian}{одно не пойму, вы все реально думаете, что они читают что то в ок или в вк?} \newline \scriptsize{(What I can't understand - do you all really think they read anything in Odnoklassniki or Vkontakte?)} \\ [0.1cm]
        detoxGPT-small & \foreignlanguage{russian}{вы читали его произведения? не читали.} \newline \scriptsize{(Have you read any of his works? No, you haven't)} \\ [0.1cm]
        detoxGPT-medium & \foreignlanguage{russian}{его посадить, извините, но мне вас очень жаль, если для вас оскорбления - норма.} \newline \scriptsize{(Putting him in jail, excuse me, but I feel sorry for you if insult is normal for you.)}\\ [0.1cm]
        detoxGPT-large & \foreignlanguage{russian}{вы читали его произведения? недалекие. попробуйте почитать посты его друзей} \newline \scriptsize{(Have you read any of his works? You are narrow-minded. Try reading his friends' posts.)}\\ [0.1cm]
        condBERT & \foreignlanguage{russian}{\underline{\smash{что ты читала из его произведений ? пусть . а если не читала , то прочти .}}} \newline \scriptsize{(Which of his works have you read? Okay. If you haven't read any, go read.)}\\ [0.1cm]
        \hline 
        
        Input & \foreignlanguage{russian}{тебя вношу в список долбаебов.} \newline \scriptsize{(I'm adding you to the list of buttheads.)}\\ [0.1cm]
        Delete & \foreignlanguage{russian}{тебя вношу в список} \newline \scriptsize{(I'm adding you to a list)}\\ [0.1cm]
        Retrieve & \foreignlanguage{russian}{надо обнародовать этот список. чего молчать.} \newline \scriptsize{(We should make this list public. Why keeping silence?)}\\ [0.1cm]
        detoxGPT-small & \foreignlanguage{russian}{Вас вношу в список людей, нуждающихся в улучшении.} \newline \scriptsize{(I'm adding you to the list of people who need being improved.)}\\ [0.1cm]
        detoxGPT-medium & \foreignlanguage{russian}{Вас вношу в список людей, которые вносят вклад в мой тред.} \newline \scriptsize{(I'm adding you to the list of people who contribute to this thread.)}\\ [0.1cm]
        detoxGPT-large & \foreignlanguage{russian}{\underline{\smash{Вас вношу в список людей, которые не соответствуют вашим ожиданиям.}}} \newline \scriptsize{(I'm adding you to the list of people who don't meet your expectations.)}\\ [0.1cm]
        condBERT & \foreignlanguage{russian}{тебя вношу в список до} \newline \scriptsize{(I'm adding you to the list of to.)}\\ [0.1cm]
        \hline
        
    \end{tabular}
    \caption{Examples of Russian texts detoxification by proposed approaches. For detoxGPT models, the results of fine-tuned setup are presented. For condBERT model, the results of Geotrend fine-tuned setup are presented. The rude words used in sentences have no goal to abuse the reader, they are just an illustration of real-life toxic texts. The best outputs for each example according to a human judgment are \underline{\smash{underlined}}.}
    \label{tab:examples}
\end{table}

Table \ref{tab:examples} shows the example outputs of the proposed models and the baselines. All the examples by detoxGPT and condBERT models were generated via the \textit{fine-tuned} scenario. The examples demonstrate the trends described above: condBERT sometimes makes an inappropriate replacement, and detoxGPT tends to output sentences not related to the input. Nevertheless, in most cases, at least one of the detoxGPT models provides a sensible answer. Interestingly, although detoxGPT-large performs best according to the metrics, the manual analysis shows that its superiority is not always evident. 

\section{Conclusion}
We presented the first study of text detoxification for the Russian language. We conducted experiments with detoxification methods based on different principles: (i)~detoxGPT model is trained on a parallel corpus and rewrites the sentence, and (ii)~condBERT is trained on non-parallel data and replaces individual toxic words with non-toxic synonyms. We described the evaluation setup, which includes the training and test data and the evaluation metrics. We evaluated the proposed methods and compare them to three simple baselines. 

The best aggregated score is achieved by detoxGPT. While condBERT shows the highest style transfer accuracy, it performs worse in naturalness preservation. However, for both methods, there is room for improvement. The detoxGPT-based models could benefit from a larger parallel corpus and more careful tuning of hyperparameters, while for condBERT, more advanced word selection strategies can increase the quality.

As a result, there is no single method that outperforms others according to all parameters of the evaluation. Sometimes it is enough to delete obscene words from the text, whereas in other cases, they should be replaced with their non-toxic synonyms. Finally, some texts can be detoxified only if fully reformulated. Thus, the most promising direction of future work would be to combine all presented strategies and apply them based on the nature of toxicity in particular sentences. 

We provide all code and data used for training and evaluation online.\footnote{https://github.com/skoltech-nlp/rudetoxifier}

\section*{Acknowledgements}

This work was conducted under the framework of the joint Skoltech-MTS laboratory. We are grateful to the anonymous reviewers for their helpful suggestions. Besides, we thank Alexey Shevtsov and Alexander Nevarko who conducted the first version of experiments with ruGPT as a part of their Deep Learning course final project at Skoltech.

\bibliography{dialogue.bib}
\bibliographystyle{plain}



\end{document}